\documentclass[runningheads]{llncs}

\usepackage{graphicx}
\usepackage{amsmath,amssymb} 
\usepackage{color}

\usepackage{epsfig}
\usepackage{graphicx}
\usepackage{bm}
\usepackage{makecell}
\usepackage{multirow}
\usepackage{comment}
\usepackage{rotating}
\usepackage{overpic}

\usepackage[
   pagebackref=false,
   breaklinks=true,
   letterpaper=true,
   colorlinks,
   bookmarks=false]{hyperref}

\def\eg{\emph{e.g.}}

\def\vs{\emph{vs.}}
\def\etal{\emph{et al.}}
\def\Hline{\Xhline{2\arrayrulewidth}}     
\newcommand{\sres} {SparseNet[$+$] }

\newcommand{\sden} {SparseNet[$\oplus$] }
\newcommand{\sdens}{SparseNets[$\oplus$] }

\newif\ifarxiv
\arxivtrue

\begin{document}

\pagestyle{headings}
\mainmatter

\title{Sparsely Aggregated Convolutional Networks}

\titlerunning{Sparsely Aggregated Convolutional Networks}

\authorrunning{L.~Zhu~\etal}

\author{%
   Ligeng Zhu$^1$
   Ruizhi Deng$^1$
   Michael Maire$^2$
   Zhiwei Deng$^1$
   Greg Mori$^1$
   Ping Tan$^1$%
}

\institute{%
   $^1$Simon Fraser University\quad%
   $^2$University of Chicago\\%
   \email{%
      \{lykenz,ruizhid,zhiweid,mori,pingtan\}@sfu.ca,
      mmaire@uchicago.edu%
   }%
}

\maketitle

\begin{abstract}
We explore a key architectural aspect of deep convolutional neural networks: the pattern of internal skip connections used to aggregate outputs of earlier layers for consumption by deeper layers.  Such aggregation is critical to facilitate training of very deep networks in an end-to-end manner.  This is a primary reason for the widespread adoption of residual networks, which aggregate outputs via cumulative summation.  While subsequent works investigate alternative aggregation operations (\eg~concatenation), we focus on an orthogonal question: which outputs to aggregate at a particular point in the network.  We propose a new internal connection structure which aggregates only a sparse set of previous outputs at any given depth.  Our experiments demonstrate this simple design change offers superior performance with fewer parameters and lower computational requirements.  Moreover, we show that sparse aggregation allows networks to scale more robustly to 1000+ layers, thereby opening future avenues for training long-running visual processes.

\end{abstract}

\section{Introduction}
\label{sec:introduction}

As convolutional neural networks have become a central component of many vision
systems, the field has quickly adopted successive improvements in their basic
design.  This is exemplified by a series of popular CNN architectures, most
notably:
AlexNet~\cite{krizhevsky2012imagenet},
VGG~\cite{simonyan2014very},
Inception~\cite{szegedy2015going,szegedy2017inception},
ResNet~\cite{he2016deep,he2016preact}, and
DenseNet~\cite{huang2016densely}.
Though initially targeted to image classification, each of these designs also
serves the role of a backbone across a broader range of vision tasks, including
object detection~\cite{girshick2014rich,maskrcnn} and
semantic segmentation~\cite{FCN,DeepLab,PSPNet}.
Advances in backbone network architecture consistently translate into
corresponding performance boosts to these downstream tasks.

We examine a core design element, internal aggregation links, of the recent
residual (ResNet~\cite{he2016deep}) and dense
(DenseNet~\cite{huang2016densely}) network architectures.  Though vital to the
success of these architectures, we demonstrate that the specific structure of
aggregation in current networks is at a suboptimal design point.  DenseNet,
considered state-of-the-art, actually wastes capacity by allocating too many
parameters and too much computation along internal aggregation links.

We suggest a principled alternative design for internal aggregation structure,
applicable to both ResNets and DenseNets.  Our design is a sparsification of
the default aggregation structure.  In both ResNet and DenseNet, the input to
a particular layer is formed by aggregating the output of all previous layers.
We switch from this full aggregation topology to one in which only a subset of
previous outputs are linked into a subsequent layer.  By changing the number of
incoming links to be logarithmic, rather than linear, in the overall depth
of the network, we fundamentally reduce the growth of parameters in our
resulting analogue of DenseNet.  Experiments reveal our design to be uniformly
advantageous:
\begin{itemize}
   \item{
      On standard tasks, such as image classification, SparseNet, our
      sparsified DenseNet variant, is more efficient than both ResNet and
      DenseNet.  This holds for measuring efficiency in terms of both
      parameters and operations (FLOPs) required for a given level of
      accuracy.  A much smaller SparseNet model matches the performance of the
      highest accuracy DenseNet.%
   }%
   \vspace{0.01\linewidth}
   \item{
      In comparison to DenseNet, the SparseNet design scales in a robust
      manner to instantiation of extremely deep networks of 1000 layers and
      beyond.  Such configurations magnify the efficiency gap between DenseNet
      and SparseNet.%
   }%
   \vspace{0.01\linewidth}
   \item{
      Our aggregation pattern is equally applicable to ResNet.  Switching
      ResNet to our design preserves or improves ResNet's performance
      properties.  This suggests that aggregation topology is a fundamental
      consideration of its own, decoupled from other design differences
      between ResNet and DenseNet.
   }
\end{itemize}

Section~\ref{sec:experiments} provides full details on these experimental
results. Prior to that, Section~\ref{sec:related} relates background on the
history and role of skip or aggregation links in convolutional neural
networks.  It places our contribution in the context of much of the recent
research focus on CNN architecture.

Section~\ref{sec:aggregation} presents the details of our sparse aggregation
strategy.  Our approach occupies a previously unexplored position in
aggregation complexity between that of standard CNNs and
FractalNet~\cite{FractalNet} on one side, and ResNet and DenseNet on the other.
Taken together with our experimental results, sparse aggregation appears to be
a simple, general improvement that is likely to filter into the standard CNN
backbone design.  Section~\ref{sec:conclusion} concludes with a synthesis of
these observations, and discussion of potential future research paths.

\section{Related Work}
\label{sec:related}

Modern CNN architectures usually consist of a series of convolutional, ReLU,
and batch normalization~\cite{ioffe2015batch} operations, mixed with
occasional max-pooling and subsampling stages.  Much prior research focuses
on optimizing for parameter efficiency within convolution, for example, via
dimensionality reduction bottlenecks~\cite{SqueezeNet,he2016deep,he2016preact},
grouped convolution~\cite{xie2016aggregated,CondenseNet}, or
weight compression~\cite{ChenWTWC15}.  These efforts all concern design at a
micro-architectural level, optimizing structure that fits inside a single
functional unit containing at most a few operations.

At a macro-architectural level, skip connections have emerged as a common and
useful design motif.  Such connections route outputs of earlier CNN layers
directly to the input of far deeper layers, skipping over the sequence of
intermediate layers.  Some deeper layers thus take input from multiple paths:
the usual sequential path as well as these shortcut paths.  Multiple intuitions
motivate inclusion of skip connections, and may share in explaining their
effectiveness.

\subsection{Skip Connections for Features}
\label{sec:related_skip}

Predicting a detailed labeling of a visual scene may require understanding it
at multiple levels of abstraction, from edges and textures to object
categories.  Taking the plausible view that a CNN learns to compute
increasingly abstract visual representations when going from shallower to
deeper layers, skip connections can provide a pathway for assembling
features that combine many levels of abstraction.  Building in such connections
alleviates the burden of learning to store and maintain features computed early
that the network needs again later.

This intuition motivates the skip connection structures found in many semantic
segmentation CNNs.  Fully convolutional networks~\cite{FCN} upsample
and combine several layers of a standard CNN, to act as input to a final
prediction layer.  Hypercolumn networks~\cite{HAGM:CVPR:2015} similarly wire
intermediate representations into a concatenated feature descriptor.  Rather
than use the end layer as the sole destination for skip links, encoder-decoder
architectures, such as SegNet~\cite{SegNet} and U-Net~\cite{UNet}, introduce
internal skip links between encoder and decoder layers of corresponding spatial
resolutions.  Such internal feature aggregation, though with different
connectivity, may also serve to make very deep networks trainable.

\subsection{Training Very Deep Networks}

Training deep networks end-to-end via stochastic gradient descent requires
backpropagating a signal through the entire network.  Starting from random
initialization, the gradient received by earlier layers from a loss at the end
of the network will be noisier than that received by deeper layers.  This issue
worsens with deeper networks, making them harder to train.  Attaching
additional losses to intermediate
layers~\cite{lee2014deeply,szegedy2015going} is one strategy for ameliorating
this problem.

Highway networks~\cite{srivastava2015highway} and residual
networks~\cite{he2016deep} (ResNets) offer a more elegant solution, preserving
the ability to train from a single loss by adding skip connections to the
network architecture.  The addition of skip connections shortens the effective
path length between early network layers and an informative loss.  Highway
networks add a gating mechanism, while residual networks implement skip
connections by summing outputs of all previous layers.  The effectiveness of
the later strategy is cause for its current widespread adoption.

Fractal networks~\cite{FractalNet} demonstrate an alternative skip connection
structure for training very deep networks.  The accompanying analysis reveals
skip connections function as a kind of scaffold that supports the training
processes.  Under special conditions, the FractalNet skip connections can be
discarded after training.

DenseNets~\cite{huang2016densely} build directly on ResNets, by switching the
operational form of skip connections from summation to concatenation.
They maintain the same aggregation topology as ResNets, as all previous layer
outputs are concatenated.

\subsection{Architecture Search}

The dual motivations of building robust representations and enabling end-to-end
training drive inclusion of internal aggregation links, but do not dictate an
optimal procedure for doing so.  Absent insight into optimal design methods,
one can treat architectural details as hyperparameters over which to
optimize~\cite{NAS}.  Training of a single network can then be wrapped as
a step in larger search procedure that varies network design.

However, it is unclear whether skip link topology is an important
hyperparameter over which to search.  Our proposed aggregation topology is
motivated by a simple construction and, as shown in
Section~\ref{sec:experiments}, significantly outperforms prior hand-designed
structures.  Perhaps our topology is near optimal and will free architecture
search to focus on more consequential hyperparameters.

\subsection{Concurrent Work}
\label{sec:related_concurrent}

Concurrent work~\cite{logDenseNet}, independent of our own, proposes a
modification of DenseNet similar to our SparseNet design.  We make distinct
contributions in comparison:
\begin{itemize}
   \item{
      Our SparseNet image classification results are substantially better than
      those reported in Hu~\etal~\cite{logDenseNet}.  Our results represent an
      actual and significant improvement over the DenseNet baseline.
   }
   \vspace{0.01\linewidth}
   \item{
      We explore sparse aggregation topologies more generally, showing
      application to ResNet and DenseNet, whereas~\cite{logDenseNet} proposes
      specific changes to DenseNet.%
   }%
   \vspace{0.01\linewidth}
   \item{
      We experiment with networks in extreme configurations (\eg~1000 layers)
      in order to highlight the robustness of our design principles in regimes
      where current baselines begin to break down.
   }
\end{itemize}

While we focus on skip connections in the context of both parameter efficiency
and network trainability, other concurrent work examines alternative mechanisms
to ensure trainability.  Xiao~\etal~\cite{DynamicalIsometry} develop a novel
initialization scheme that allows training very deep vanilla CNNs.
Chang~\etal~\cite{RevArbDeep}, taking inspiration from ordinary differential
equations, develop a framework for analyzing stability of reversible
networks~\cite{RevNet} and demonstrate very deep reversible architectures.

\section{Aggregation Architectures}
\label{sec:aggregation}

\begin{figure*}
   \begin{center}
      \includegraphics{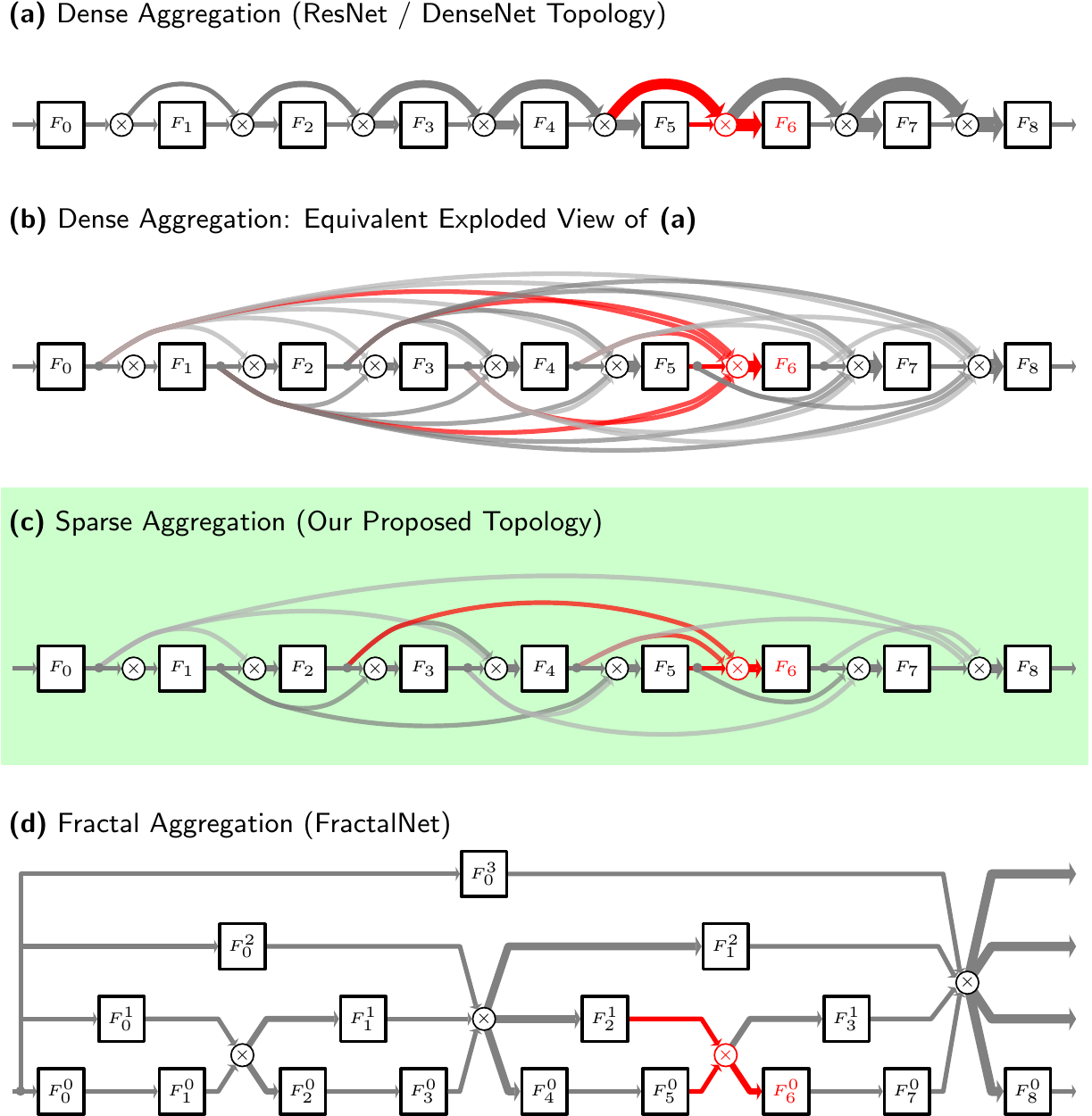}
   \end{center}
   \caption{
      \textbf{Aggregation topologies.}
      Our proposed sparse aggregation topology devotes less machinery to
      skip connections than DenseNet~\cite{huang2016densely}, but more than
      FractalNet~\cite{FractalNet}.  This is apparent by comparing the exploded
      view (b) of the ResNet~\cite{he2016deep} or DenseNet topology (a), as
      well as the fractal topology (d), with our proposal (c).  All of these
      architectures are describable in terms of a basic parameterized
      functional unit $F(\cdot)$ (\eg~convolution-ReLU-batchnorm), an
      aggregation operator $\otimes$, and a connection pattern.  For ResNet,
      $\otimes$ is addition [$+$]; for DenseNet, $\otimes$ is concatenation
      [$\oplus$]; for FractalNet $\otimes$ is averaging [$\overline{+}$].
      Note how the compact view (a) feeds the result of one aggregation into
      the next; the exploded view (b) of DenseNet is the correct visualization
      for comparison to (c) and (d).  For a network of depth $N$, dense
      aggregation requires $O(N^2)$ connections, sparse aggregation
      $O(N\log(N))$, and fractal aggregation $O(2N)$.  These differences are
      visually apparent by comparing incoming links at a common depth.  For
      example, compare the density of the (highlighted red) links into
      layer~$F_6(\cdot)$.
   }
   \label{fig:aggregation_topology}
\end{figure*}

Figure~\ref{fig:aggregation_topology} sketches our proposed sparse aggregation
architecture alongside the dominant ResNet~\cite{he2016deep} and
DenseNet~\cite{huang2016densely} designs, as well as the previously proposed
FractalNet~\cite{FractalNet} alternative to ResNet.  This macro-architectural
view abstracts away details such as the specific functional unit $F(\cdot)$,
parameter counts and feature dimensionality, and the aggregation
operator~$\otimes$.  As our focus is on a novel aggregation topology,
experiments in Section~\ref{sec:experiments} match these other details to those
of ResNet and DenseNet baselines.

We define a network with a sparse aggregation structure to be a sequence of
nonlinear functional units (layers) $F_{\ell}(\cdot)$ operating on input $x$,
with the output $y_{\ell}$ of layer $\ell$ computed as:
\begin{eqnarray}
   &y_0      = F_0(x) \\
   &y_{\ell} = F_{\ell}(
      \otimes(
         y_{\ell - c^{0}},
         y_{\ell - c^{1}},
         y_{\ell - c^{2}},
         y_{\ell - c^{3}},
         \ldots,
         y_{\ell - c^{k}}))
\end{eqnarray}
where $c$ is a positive integer and $k$ is the largest non-negative integer
such that $c^k \le \ell$.  $\otimes$ is the aggregation function.  This amounts
to connecting each layer to previous layers at exponentially increasing
offsets.  Contrast with ResNet and DenseNet, which connect each layer to
all previous layers according to:
\begin{eqnarray}
   y_{\ell} = F_{\ell}(
      \otimes(
         y_{\ell - 1},
         y_{\ell - 2},
         y_{\ell - 3},
         y_{\ell - 4},
         \ldots,
         y_{0}))
\end{eqnarray}
For a network of total depth $N$, the full aggregation strategy of ResNet and
DenseNet introduces $N$ incoming links per layer, for a total of $O(N^2)$
connections.  In contrast, sparse aggregation introduces no more than
$\log_c(N)$ incoming links per layer, for a total of $O(N\log(N))$ connections.

Our sparse aggregation strategy also differs from FractalNet's aggregation
pattern.  The FractalNet~\cite{FractalNet} design places a network of depth
$N$ in parallel with networks of depth $\frac{N}{2}$, $\frac{N}{4}$, \ldots,
$1$, making the total network consist of $2N-1$ layers.  It inserts occasional
join (aggregation) operations between these parallel networks, but does so
with such extreme sparsity that the total connection count is still dominated
by the $O(2N)$ connections in parallel layers.

Our sparse connection pattern is sparser than ResNet or DenseNet, yet denser
than FractalNet.  It occupies a previously unexplored point, with a
fundamentally different scaling rate of skip connection density with network
depth.

\subsection{Potential Drawbacks of Dense Aggregation}
\label{sec:aggregation_dense_drawbacks}

The ability to train networks with depth greater than $100$ layers using
DenseNet and ResNet architectures can be partially attributed to to their
feature aggregation strategies.  As discussed in Section~\ref{sec:related},
skip links serve as a training scaffold, allowing each layer to be directly
supervised by the final output layer, and aggregation may help transfer useful
features from shallower to deeper layers.

However, dense feature aggregation comes with several potential drawbacks.
These drawbacks appear in different forms in the ResNet-styled aggregation by
summation and the DenseNet-styled aggregation by concatenation, but share a
common theme of over-constraining or over-burdening the system.

In general, it is impossible to disentangle the original components of a set of
features after taking their sum.  As the depth of a residual network grows,
the number of features maps aggregated grows linearly.  Later features may
corrupt or wash-out the information carried by earlier feature maps.  This
information loss caused by summation could partially explain the saturation of
ResNet performance when the depth exceeds $1000$ layers~\cite{he2016deep}.
This way of combining features is also hard-coded in the design of ResNets,
giving the model little flexibility to learn more expressive combination
strategies.  This constraint may be the reason that ResNet layers tend to
learn to perform incremental feature updates~\cite{unrolled}.

In contrast, the aggregation style of DenseNets combines features through
direct concatenation, which preserves the original form of the previous
features.  Concatenation allows each subsequent layer a clean view of all
previously computed features, making feature reuse trivial.  This factor may
contribute to the better parameter-performance efficiency of DenseNet over
ResNet.

But DenseNet's aggregation by concatenation has its own problems: the number of
skip connections and required parameters grows at a rate of $O(N^2)$, where $N$
is the network depth.  This asymptotically quadratic growth means that a
significant portion of the network is devoted to processing previously seen
feature representations.  Each layer contributes only a few new outputs to an
ever-widening concatenation of stored state.  Experiments show that it is hard
for the model to make full use of all the parameters and dense skip
connections.  In the original DenseNet work~\cite{huang2016densely}, a large
fraction of the skip connections have average absolute weights of convolution
filters close to zero.  This implies that dense aggregation of feature maps
maintains some extraneous state.

The pitfalls of dense feature aggregation in both DenseNet and ResNet are
caused by the linear growth in the number of features aggregated with respect
to the depth.  Variants of ResNet and DenseNet, including the
post-activation ResNets~\cite{he2016preact}, mixed-link
networks~\cite{MixedLink}, and dual-path networks~\cite{DualPath} all use the
same dense aggregation pattern, differing only by aggregation operator.  They
thus inherit potential limitations of this dense aggregation topology.

\subsection{Properties of Sparse Aggregation}
\label{sec:aggregation_sparse_props}

\begin{table}[b]
   \begin{center}
   \begin{tabular}{c|c|c|c}
                          & Parameters     & Shortest Gradient Path & Aggregated Features\\ \Hline
      Plain               &  $O(N)$        &    $O(N)$              &         $O(1)$     \\ \hline
      ResNets             &  $O(N)$        &    $O(1)$              &         $O(\ell)$  \\ \hline
      DenseNets           &  $O(N^2)$      &    $O(1)$              &         $O(\ell)$  \\ \hline
      SparseNets (sum)    &  $O(N)$        &    $O(\log(N))$        &   $O(\log{\ell})$  \\ \hline
      SparseNets (concat) &  $O(N\log{N})$ &    $O(\log(N))$        &   $O(\log{\ell})$  \\ \hline
   \end{tabular}
   \end{center}
   \caption{
      \textbf{SparseNet properties.} We compare architecture-induced scaling
      properties for networks of depth $N$ and for individual layers located at
      depth $\ell$.
   }
   \label{tab:sparsenet_props}
\end{table}

We would like to maintain the power of short gradient paths for training,
while avoiding the potential drawbacks of dense feature aggregation.
SparseNets do, in fact, have shorter gradient paths than architectures without
aggregation.

In plain feed-forward networks, there is only one path from a layer to a
previous layer with offset $S$; the length of the path is $O(S)$.  The
length of the shortest gradient path is constant in dense aggregation networks
like ResNet and DenseNet.  However, the cost of maintaining a gradient path
with $O(1)$ length between any two layers is the linear growth of the count
of aggregated features.  By aggregating features only from layers with
exponential offset, the length of the shortest gradient path between two layers
with offset $S$ is bounded by $O((c-1)\log(S))$.  Here, $c$ is again the base
of the exponent governing the sparse connection pattern.

It is also worth noting that the number of predecessor outputs gathered by the
$\ell^{\textrm{th}}$ layer is $O(\log(\ell))$, as it only reaches predecessors
with exponential offsets.  Therefore, the total number of skip connections is
\begin{equation}
    \sum_{\ell=1}^{N} \lfloor \log_{c}{\ell} \rfloor = O(N \log{N})
    \label{bound_total_connections}
\end{equation} where $N$ is the number of layers (depth) of the network.
The number of parameters are $O(N \log{N})$ and $O(N)$, respectively, for
aggregation by concatenation and aggregation by summation.
Table~\ref{tab:sparsenet_props} summarizes these properties.

\section{Experiments}
\label{sec:experiments}

We demonstrate the effectiveness SparseNets as a drop-in replacement (and
upgrade) for state-of-the-art networks with dense feature aggregation, namely
ResNets~\cite{he2016deep,he2016preact} and DenseNets~\cite{huang2016densely},
through image classification tasks on the CIFAR~\cite{krizhevsky2009learning}
and ImageNet datasets~\cite{deng2009imagenet}.  Except for the difference
between the dense and sparse aggregation topologies, we set all other
SparseNet hyperparameters to be the same as the corresponding ResNet or
DenseNet baseline.

For some large models, image classification accuracy appears to saturate when
we continue increasing model depth or internal channel counts.  It is likely
such saturation is not due to model capacity limits, but rather both our
model and baselines reach diminishing returns given the dataset size and
task complexity.  We are interested not only in absolute accuracy, but also
parameter-accuracy and FLOP-accuracy efficiency.

We implement our models in the TensorFlow framework~\cite{tensorflow}.  For
optimization, we use SGD with Nesterov momentum 0.9 and weight decay 0.0001.
We train all models from scratch using He~\etal's initialization
scheme~\cite{he2015delving}.  All networks were trained using NVIDIA
GTX 1080 Ti GPUs. We release our implementation\footnote{%
\url{https://github.com/Lyken17/SparseNet}} of SparseNets, with full
details of model architecture and parameter settings, for the purpose of
reproducible experimental results.

\subsection{Datasets}

\subsubsection{CIFAR}
Both the CIFAR-10 and CIFAR-100 datasets~\cite{krizhevsky2009learning} have
50,000 training images and 10,000 testing images with size of $32 \times 32$
pixels. CIFAR-10 (C10) and CIFAR-100 (C100) have 10 and 100 classes
respectively.  Our experiments use standard data augmentation, including
mirroring and shifting, as done in~\cite{huang2016densely}.  The mark $+$
beside C10 or C100 in results tables indicates this data augmentation scheme.
As preprocessing, we normalize the data by the channel mean and standard
deviation.  Following the schedule from the Torch implementation of
ResNet~\cite{fb_resnet_torch},
our learning rate starts from 0.1 and is divided by 10 at epoch 150 and 225.

\subsubsection{ImageNet}
The ILSVRC 2012 classification dataset~\cite{deng2009imagenet} contains 1.2
million images for training, and 50K for validation, from 1000 classes.  For a
fair comparison, we adopt the standard augmentation scheme for training images
as in~\cite{fb_resnet_torch,huang2016densely,he2016deep}.
Following~\cite{he2016deep,huang2016densely}, we report classification errors
on the validation set with single crop of size 224 $\times$ 224.

\subsection{Results on CIFAR}

Table~\ref{tab:cifar_results} reports experimental results on
CIFAR~\cite{krizhevsky2009learning}.  The best SparseNet closely matches the
performance of the state-of-the DenseNet.
In all of these experiments, we instantiate each SparseNet to be exactly the same as the
correspondingly named DenseNet, but with sparser aggregation structure (some
connections removed).  The parameter $k$ indicates feature growth rates (how
many new feature channels each layer produces), which we match to the DenseNet
baseline.  Models whose names end with BC use the bottleneck compression
structure, as in the original DenseNet paper.  As SparseNet does fewer
concatenations than DenseNet, the same feature growth rate produces models with
fewer overall parameters.  Remarkably, for many of the corresponding $100$
layer models, SparseNet performs as well or better than DenseNet, while having
substantially fewer parameters.

\begin{table*}[bh!]
   \scriptsize
   \centering
   \begin{tabular}[t]{l|cc|c|cc}
      Architecture & Depth & Params & C10+ & C100+ \\
      \hline
      ResNet \cite{he2016deep} & 110 & 1.7M & 6.61 & - \\
      ResNet(pre-activation)\cite{he2016deep} & 164 &   1.7M & 5.46 & 24.33  \\
      ResNet(pre-activation)\cite{he2016deep} & 1001 & 10.2M & 4.62 & 21.42* \\
      \hline
      Wide ResNet \cite{sergey2016wideresnet} & 16 & 11.0M &  4.81 & 22.07 \\
      FractalNet \cite{FractalNet}            & 21 &    38.6M &  5.52 & 23.30 \\
      \hline
      DenseNet (k=12)\cite{huang2016densely}  &  40 &  1.1M   & 5.39* & 24.79* \\
      DenseNet (k=12)\cite{huang2016densely}  & 100 &  7.2M   & 4.28* & 20.97* \\
      DenseNet (k=24)\cite{huang2016densely}  & 100 & 28.3M  & 4.04* & 19.61* \\
      DenseNet (k=16, 32, 64)\cite{huang2016densely} & 100 & 61.1M  & 4.31* & 20.6* \\
      DenseNet (k=32, 64, 128)\cite{huang2016densely} & 100 & 241.6M  & N/A & N/A \\
      DenseNet-BC (k=24)\cite{huang2016densely}  & 250 &  15.3M  & \textbf{3.65}  & 17.6 \\
      DenseNet-BC (k=40)\cite{huang2016densely}  & 190 &  25.6M  & 3.75*  &  \textbf{17.53*} \\
      DenseNet-BC (k=16, 32, 64)\cite{huang2016densely}  & 100 &  7.9M  & 4.02*  &  19.55* \\
      DenseNet-BC (k=32, 64,128)\cite{huang2016densely}  & 100 &  30.5M  & 3.92*  &  18.71* \\
      \hline
      SparseNet (k=12)          &  40 & 0.8M  & 5.47  & 24.48  \\
      SparseNet (k=24)          & 100 & 2.5M  & 4.44  & 22.73  \\
      SparseNet (k=36)          & 100 & 5.7M  & 4.34  & 21.10  \\
      SparseNet (k=16, 32, 64)  & 100 & 7.2M  & 4.21 & 19.89   \\
      SparseNet (k=32, 64, 128) & 100 & 27.7M & 3.99 & 19.30   \\
      SparseNet-BC (k=24)       & 100 & 1.5M  & 4.49  & 22.71  \\
      SparseNet-BC (k=36)       & 100 & 3.3M  & 4.27  & 21.40  \\
      SparseNet-BC (k=16, 32, 64)   & 100 &  4.4M  &  4.34  & 19.90  \\
      SparseNet-BC (k=32, 64, 128)  & 100 & 16.7M  &  \textbf{4.10}  & \textbf{18.22}  \\
      \hline
   \end{tabular}
   \vspace{0.03\linewidth}
   \caption{
      \textbf{CIFAR classification performance.}
      We show classification error rate for SparseNets compared to DenseNets,
      ResNets, and their variants.  Results marked with a $*$ are from our
      implementation.  Datasets marked with + indicates use of standard data
      augmentation (translation and mirroring).
   }
   \label{tab:cifar_results}
\end{table*}

\ifarxiv
\fi

\subsection{Going Deeper with Sparse Connections}

\begin{table}[t]
   \centering
   \begin{tabular}{cc}
      \begin{minipage}[t]{0.46\linewidth}
         \scriptsize{
            \begin{tabular}[t]{c|c c|c}
               \hline
               Model & Depth & Params & CIFAR 100+ \\
               \hline
               \multirow{5}{*}{ResNet}
                  & 56  & 0.59M & 27.00 \\ 
                  & 110 & 1.15M & 24.70 \\
                  & 200  &  2.07M & 23.10 \\
                  & 1001 & 10.33M & \textbf{21.42} \\
                  & 2000 & 20.62M & 22.76 \\
               \hline
               \multirow{5}{*}{SparseNet[$+$]}
                  &   56  &  0.59M  & 27.70  \\
                  &  110  &  1.15M  & 26.10  \\
                  &  200  &  2.07M & 25.77  \\
                  & 1001  & 10.33M &  22.10  \\
                  & 2000  & 20.62M & \textbf{ 21.01}  \\ \hline
            \end{tabular}
         }
      \end{minipage}
      &
      \begin{minipage}[t]{0.46\linewidth}
         \scriptsize{
            \begin{tabular}[t]{c|c c|c}
               \hline
               Model & Depth & Params & CIFAR 100+ \\
               \hline
               \multirow{3}{*}{DenseNet(k=12)} &   40  & 1.10M & 24.79 \\
                                               &  100  & 7.20M & 20.97 \\
                                               &  400  & 117M  &  N/A  \\
               \hline
               \multirow{2}{*}{DenseNet-BC(k=24)}
                   &  250  & 25.6M  & 17.6 \\
                   &  400  & 216.3M & N/A \\
               \hline
               \multirow{2}{*}{DenseNet-BC(k=4)}
                   &  400  & 1.10M & 32.94 \\
                   & 1001  & 6.63M & 28.50 \\
               \hline
               \multirow{2}{*}{
                       \begin{tabular}{l}
                           SparseNet[$\oplus$]-BC \\  (k=12)
                       \end{tabular}
                   }
                   &  100 &  0.40M & 27.89  \\
                   &  400 &  1.70M & 24.53  \\
               \hline
            \end{tabular}
         }
      \end{minipage}
   \end{tabular}
   \vspace{0.02\linewidth}
   \caption{
      \textbf{Depth scalability on CIFAR.}
      \textbf{\emph{Left:}}
         ResNets and their sparsely aggregated analogue SparseNets[$+$].
      \textbf{\emph{Right:}}
         DenseNets and their corresponding sparse analogues SparseNets[$\oplus$].
         Observe that ResNet and all SparseNet variants of any depth
         exhibit robust performance.  DenseNets suffer an efficiency
         drop when stretched too deep.
   }
   \label{tab:cifar_sparse_res}
\end{table}

Table~\ref{tab:cifar_sparse_res} shows results of pushing architectures to
extreme depth.  While Table~\ref{tab:cifar_results} explored only the SparseNet
analogue of DenseNet, we now explore switching both ResNet and DenseNet to
sparse aggregation structures, and denote their corresponding SparseNets by
SparseNet[$+$] and SparseNet[$\oplus$], respectively.

Both ResNet and SparseNet[$+$] demonstrate better performance on CIFAR100 as
their depth increases from 56 to 200 layers.  The gap between the performance of
ResNet and SparseNet[$+$] initially enlarges as depth increases.  However, it
narrows as network depth reaches 1001 layers, and the performance of
SparseNet[$+$]-2000 surpasses ResNet-2000.  Compared to ResNet, \sres
appears better able to scale to depths of over 1000 layers.

Similar to both ResNet and SparseNet[$+$], the performance of DenseNet and
\sden also improves as their depth increases.  The performance of
DenseNet is also affected by the feature growth rate.  However, the parameter
count of DenseNet explodes as we increase its depth to 400, even with a growth
rate of 12.  Bottleneck compression layers have to be adopted and the number of
filters in each layer has to be significantly reduced if we want to go deeper.
We experiment with DenseNets of depth greater than 1000 by adopting
bottleneck compression (BC) structure and using a growth rate of 4.  But, as
Table~\ref{tab:cifar_sparse_res} shows, their performance is far from
satisfying.  In contrast, building SparseNet[$\oplus$] to extreme depths is
practical and memory-efficient.  We can easily build SparseNet[$\oplus$] to
400 layers using BC structure and a growth rate of 12.  At such depth, DenseNet
is only practical with a restricted growth rate ($k=4$) and unable to compete
in terms of accuracy.

An important advantage of SparseNet[$\oplus$] over DenseNet is that the number
of previous layers aggregated can be bounded by a small integer even when the
depth of the network is hundreds of layers.  This is a consequence of the slow
growth rate of the logarithm function.  This feature not only permits
building deeper SparseNet[$\oplus$] variants, but allows us to explore
hyperparameters of SparseNet[$\oplus$] with more flexibility on both depth and
filter count.

We also observe that SparseNet[$\oplus$] generally has better parameter
efficiency than SparseNet[$+$].  For example, on CIFAR-100, the parameter
count of SparseNet[$+$]-200 is slightly higher than SparseNet[$\oplus$]-400.
However, notice that SparseNet[$\oplus$]-400 outperforms SparseNet[$+$]-200.
The standard DenseNet \vs~ResNet advantage of preserving features via
concatenation (over summation) may hold for the sparse aggregation pattern.

\subsection{Efficiency of SparseNet[$\oplus$]}

Returning to Table~\ref{tab:cifar_results}, we can further comment on the
efficiency of SparseNet[$\oplus$] (denoted in Table~\ref{tab:cifar_results} as
SparseNet) in comparison to DenseNet.  These results include our exploration of
parameter efficiency by varying the depth and number of filters of
SparseNet[$\oplus$].  As the number of features each layer aggregates grows
slowly, and is nearly a constant within a block, we also double the number of
filters across blocks, following the approach of ResNets.  Here, a block
refers to a sequence of layers running at the same spatial resolution, between
pooling and subsampling stages of the CNN pipeline.

There are two general trends in the results.  First, SparseNet usually requires
fewer parameters than DenseNet when they have close performance.  Looking at
the $100$-layer networks, SparseNet ($k=36$) has accuracy comparable to
DenseNet ($k=12$), using fewer parameters.  The $18.22\%$ error rate of
SparseNet-BC ($N=100$, $k=\{32, 64, 128\}$) is better than the performance of
the corresponding DenseNet-BC ($N=100$, $k=\{32, 64, 128\}$) and requires only
16.7 rather than 30.5 million parameters.

Second, when both networks have less than 15 million parameters, SparseNet
consistently outperforms the DenseNet with similar parameter count.  For
example, DenseNet ($N=100, k=12$) and SparseNet ($N=100, k=\{16,32,64\}$) both
have around 7.2 million parameters, but the latter shows better performance.
DenseNet ($N=40, k=12$) consumes around 1.1 million parameters but still has
worse performance on CIFAR-100 than the 0.8 million parameter SparseNet
($N=40, k=12$).

Counterexamples do exist, such as the comparison between
SparseNet-BC-100-\{32, 64, 128\} and DenseNet-BC-250-24.  The latter model,
with fewer parameters, performs better ($17.53\%$ vs $18.22\%$ error)
than the previous one.  However, network depth may be a conflating factor here,
as the comparison involves a DenseNet over twice as deep as the SparseNet.
Controlling for depth, SparseNet induces a different parameter allocation
strategy than DenseNet; internal layers can produce wider outputs as they are
concatenated together less frequently.

Note that when we double the number of filters across different blocks, the
performance of SparseNets is boosted and their better parameter efficiency
over DenseNets becomes more obvious.  SparseNets achieve similar or better
performance than DenseNets, while requiring about half the number of
parameters uniformly across all settings.  These general trends are
summarized in the parameter-performance plots in Figure~\ref{fig:perform_param}
(left).

\subsection{Results on ImageNet}

\begin{table}[t]
   \centering
   \small
   \begin{tabular}[t]{l|c|ccc}
      Model & Error & Params & FLOPs & Time  \\
      \hline
      DenseNet-121-32 \cite{huang2016densely} & 25.0* & 7.98M  & 5.7G & 19.5ms \\
      DenseNet-169-32 \cite{huang2016densely} & 23.6* & 14.15M & 6.76G & 32.0ms \\
      DenseNet-201-32 \cite{huang2016densely} & 22.6* & 20.01M & 8.63G & 42.6ms \\
      DenseNet-264-32 \cite{huang2016densely} & \textbf{22.2*} & 27.21M & 11.03G & 50.4ms \\
      \hline
      SparseNet[$\oplus$]-121-32 &  25.8  &  4.51M  & 3.46G & 13.5ms\\
      SparseNet[$\oplus$]-169-32 &  24.5  &  6.23M  & 3.74G & 18.8ms\\
      SparseNet[$\oplus$]-201-32 &  23.6  &  7.22M  & 4.13G & 22.0ms \\
      SparseNet[$\oplus$]-201-48 &  \textbf{22.7}   & 14.91M  & 9.19G & 43.1ms
      \\
      \hline
      ResNet-50  &  23.9  &  25.5M   & 8.20G & 42.2ms \\
      ResNet-50 Pruned~\cite{han2015deep_compress} & \textbf{23.7} & 7.47M & - & - \\
      \hline
   \end{tabular}
   \vspace{0.02\linewidth}
   \caption{
      \textbf{ImageNet results.}
      The top-1 single-crop validation error, parameters, FLOPs, and time of
      each model on ImageNet.
   }
   \label{tab:imagenet_results}
\end{table}

To demonstrate efficiency on a larger-scale dataset, we further test different
configurations of \sden and compare them with state-of-the-art networks on
ImageNet.  All models are trained with the same preprocessing methods and
hyperparameters.  Table~\ref{tab:imagenet_results} reports ImageNet validation
error.

These results reveal that the better parameter-performance efficiency exhibited
by SparseNet[$\oplus$] over DenseNet extends to ImageNet~\cite{deng2009imagenet}:
\sden performs similarly to state-of-the-art DenseNets, while requiring
significantly fewer parameters.  For example, SparseNet-201-48 (14.91M params)
yields comparable validation error to DenseNet-201-32 (20.01M params).
SparseNet-201-32 (7.22M params) matches DenseNet-169-32 with just half the
parameter count.

\begin{figure}
   \begin{center}
      \begin{minipage}[t]{0.49\linewidth}
         \centering
         \includegraphics[width=1\linewidth]{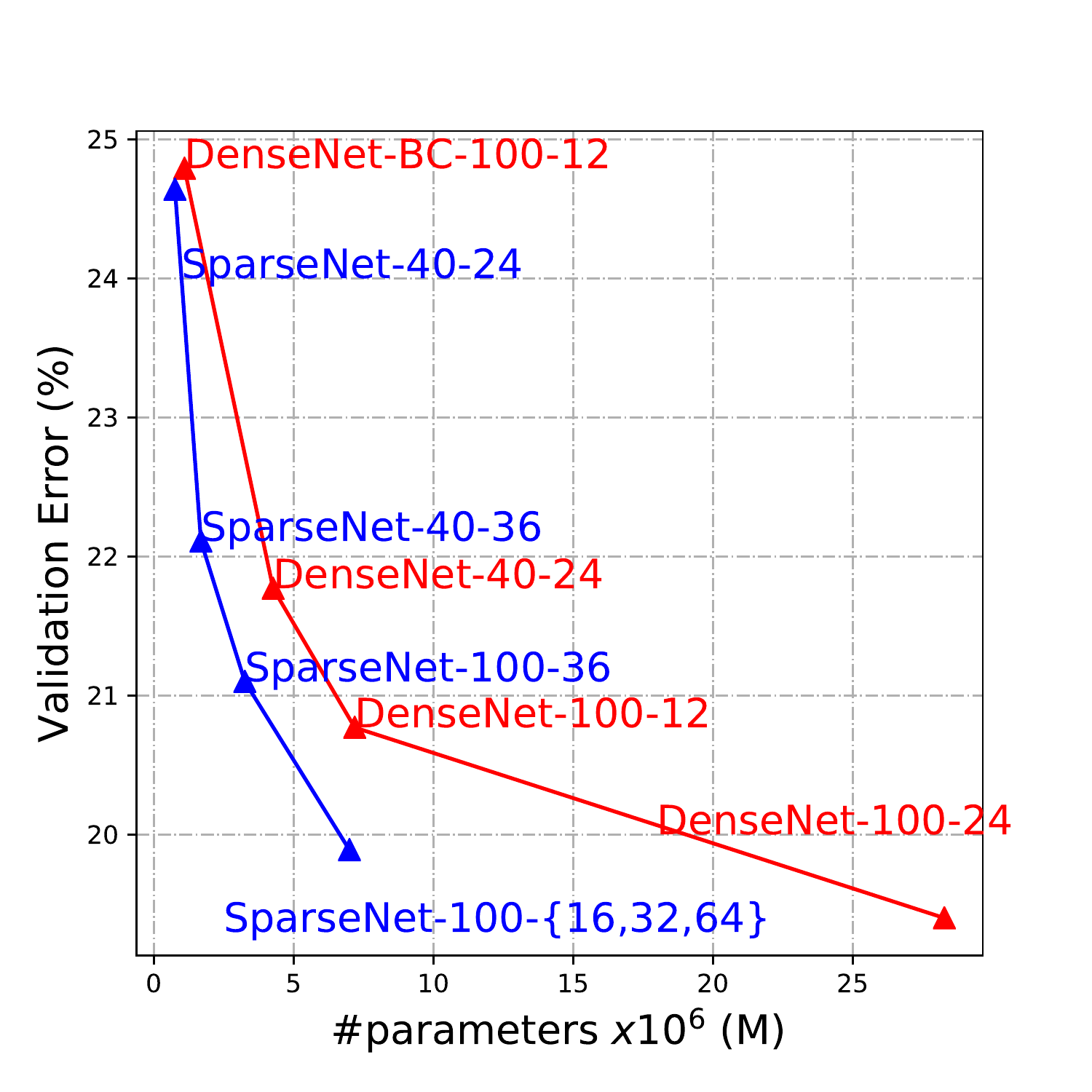}
      \end{minipage}
      \begin{minipage}[t]{0.49\linewidth}
         \centering
         \includegraphics[width=1\linewidth]{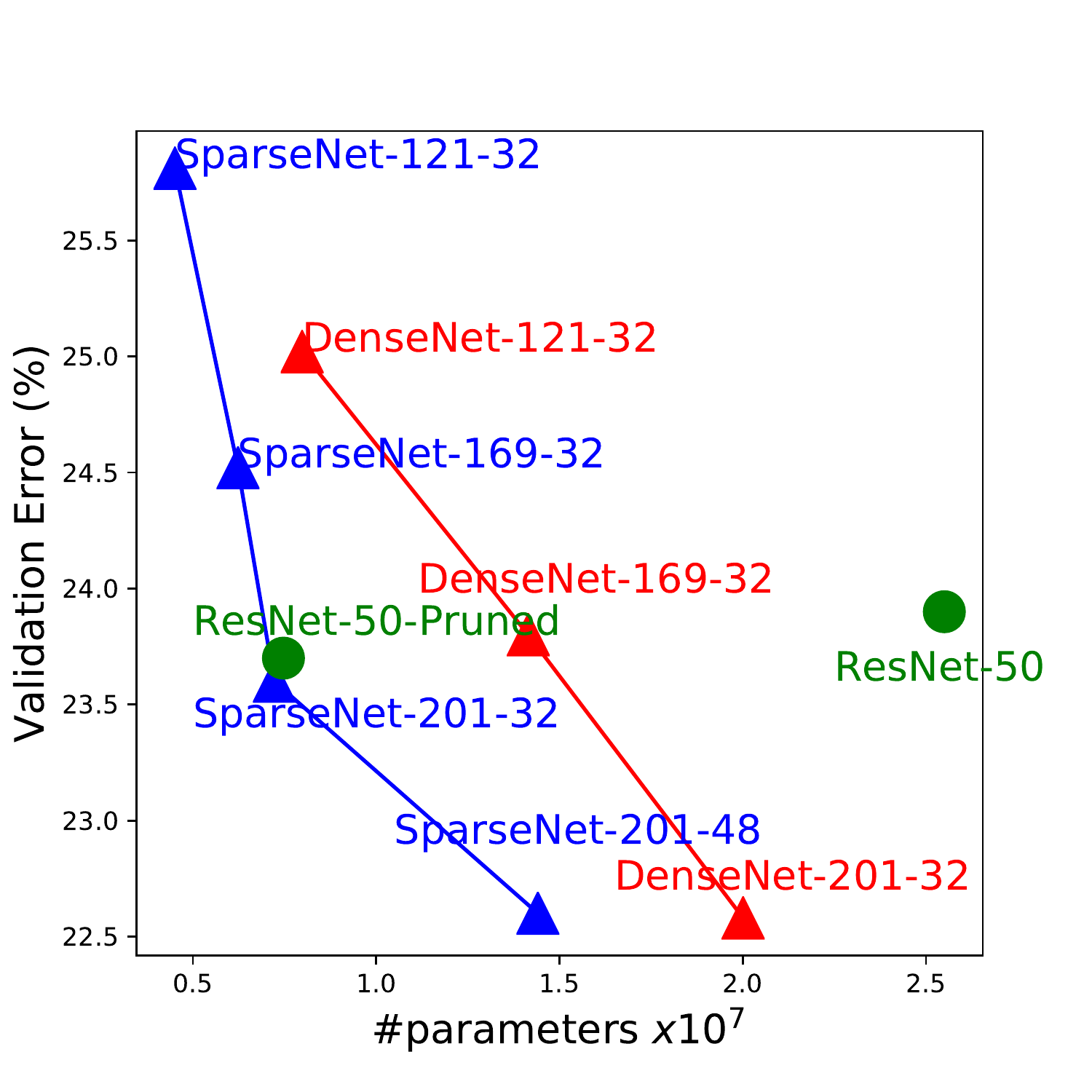}
      \end{minipage}
   \end{center}
   \caption{
      \textbf{Parameter efficiency.}
      Comparison between DenseNets and \sdens on top-1$\%$ error and number of
      parameters with different configurations.
      \textbf{\emph{Left:}} CIFAR.
      \textbf{\emph{Right:}} ImageNet.
      SparseNets achieve lower error with fewer parameters.
   }
   \label{fig:perform_param}
\end{figure}

Even compared to pruned networks, SparseNets show competitive parameter
efficiency.  In the last row of the Table~\ref{tab:imagenet_results}, we
show the result of pruning ResNet-50 using deep
compression~\cite{han2015deep_compress}, whose parameter-performance efficiency
significantly outpaces the unpruned ResNet-50.  However, our SparseNet[$\oplus$]-201-32,
trained from scratch, has even better error rate than pruned ResNet-50, with
fewer parameters.  See Figure~\ref{fig:perform_param} (right) for a complete
efficiency plot.

\subsection{Feature Reuse and Parameter Redundancy}

\begin{figure}
   \begin{center}
      \includegraphics[width=0.9\linewidth]{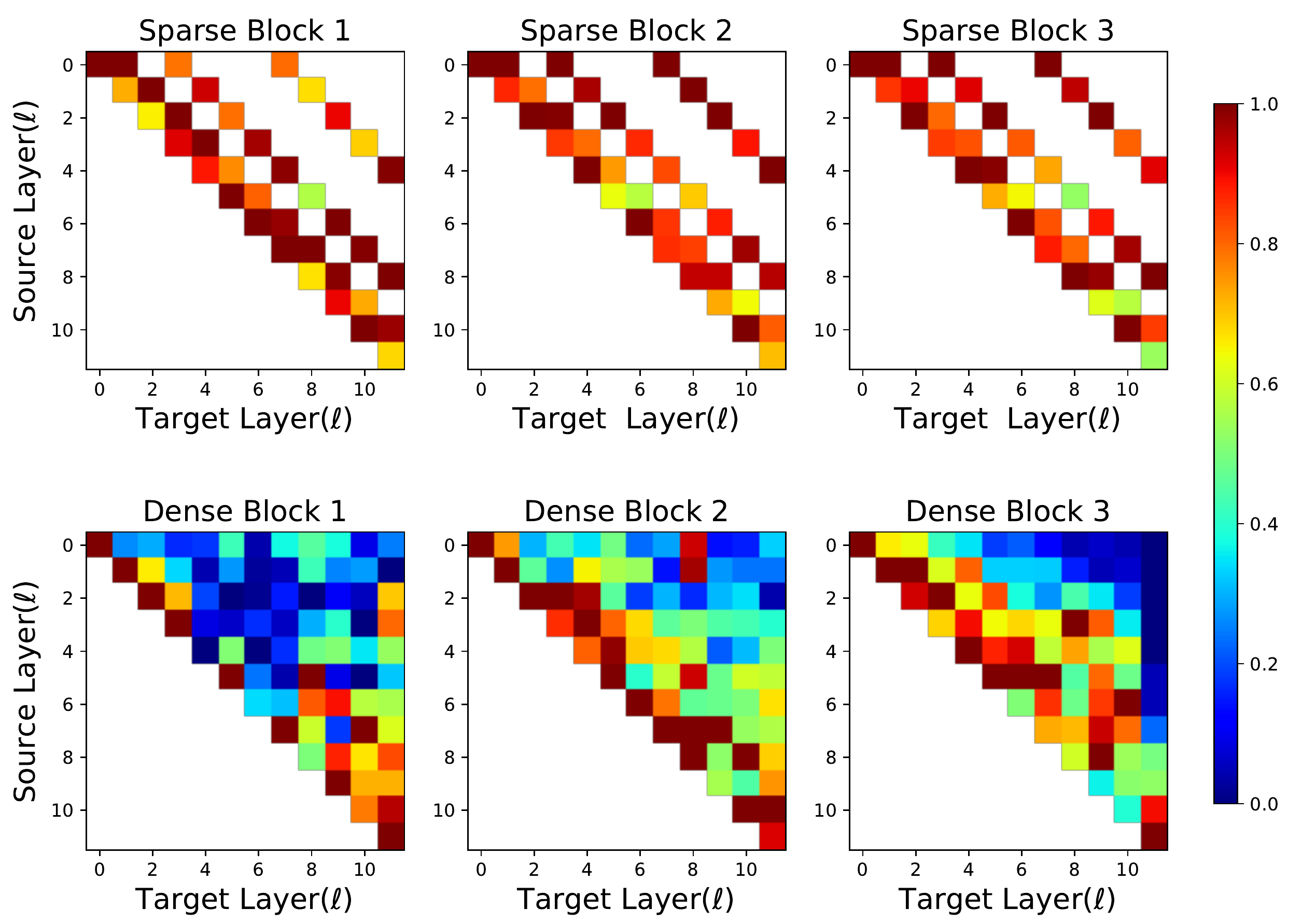}
   \end{center}
   \caption{
      The average absolute filter weights of convolutional layers in trained
      DenseNet and SparseNet.  The color of pixel $(i, j)$ indicates the
      average weights of connections from layer $i$ to $j$ within a block.
      The first row encodes the weights attached to the first input layer of a
      DenseNet/SparseNet block.
   }
   \label{fig:weight_distribution}
\end{figure}

The original DenseNets work~\cite{huang2016densely} conducts a simple
experiment to investigate how well a trained network reuses features across
layers.  In short, for each layer in each densely connected block, they
compute the average absolute weights of the part of that layer's filters that
convolves with each previous layer's feature map.  The averaged absolute
weights are rescaled between 0 and 1 for each layer $i$.  The $j^\textrm{th}$
normalized value implies the relative dependency of the features of layer $i$
on the features of layer $j$, compared to other layers.  These experiments are
performed on a DenseNet consisting of 3 blocks with $N=40$ and $k=12$.

We perform a similar experiment on a SparseNet model with the same
configuration.  We plot results as heat maps in
Figure~\ref{fig:weight_distribution}.  For comparison, we also include the heat
maps of the corresponding experiment on DenseNets~\cite{huang2016densely}.  In
these heap maps, a red pixel at location $(i, j)$ indicates layer $i$
makes heavy use of the features of layer $j$; a blue pixel indicates
relatively little usage.  A white pixel indicates there is no direct connection
between layer $i$ and layer $j$.  From the heat maps, we observe the
following:
\begin{itemize}
    \item{
      Most of the non-white elements in the heat map of SparseNet are close to
      red, indicating that each layer takes full advantage of all the features
      it directly aggregates.  It also indicates almost all the parameters are
      fully exploited, leaving little parameter redundancy.  This result is not
      surprising considering the high observed parameter-performance efficiency
      of our model.
   }
   \vspace{0.01\linewidth}
   \item{
      In general, the layer coupling value at position $(i, j)$ in DenseNet
      decreases as the offset between $i$ and $j$ gets larger.  However,
      such a decaying trend does not appear in the heat map of SparseNet,
      implying that layers in SparseNets have better ability to extract useful
      features from long-distance connections to preceding layers.
   }
\end{itemize}

The distribution of learned weights in Figure~\ref{fig:weight_distribution},
together with the efficiency curves in Figure~\ref{fig:perform_param}, serves
to highlight the importance of optimizing macro-architectural design.  Others
have demonstrated the benefits of a range of schemes
\cite{SqueezeNet,LearnStructSparse,DeepExpander,gray2017gpu,CondenseNet}
for sparsifying micro-architectural network structure (parameter
structure within layers or filters).  Our results show similar considerations
are relevant at the scale of the entire network.

\section{Conclusion}
\label{sec:conclusion}

We demonstrate that following a simple design rule, scaling aggregation link
complexity in a logarithmic manner with network depth, yields a new
state-of-the-art CNN architecture.  Extensive experiments on CIFAR and ImageNet
show our SparseNets offer significant efficiency improvements over the widely
used ResNets and DenseNets.  This increased efficiency allows SparseNets to
scale robustly to great depth.  While CNNs have recently moved from the
10-layer to 100-layer regime, perhaps new possibilities will emerge with
straightforward and robust training of 1000-layer networks.

The performance of neural networks for visual recognition has grown
with their depth as they evolved from AlexNet~\cite{krizhevsky2012imagenet} to
ResNet~\cite{he2016deep,he2016preact}.  Extrapolating such trends could be
cause to believe building deeper networks should further improve performance.
Much effort has been devoted by researchers in computer vision and machine
learning communities to train deep neural networks with more than 1000 layers,
with such hope~\cite{he2016preact,huang2016deep}.

Though previous works and our experiments show we can train very deep neural
networks with stochastic gradient descent, their test performance still usually
plateaus. Even so, very deep neural networks might be suitable for other
interesting tasks.  One possible future direction could be solving sequential
search or reasoning tasks relying on long-term dependencies.  Skip connections
might empower the network with backtracking ability.  Sparse feature
aggregation might permit building extremely deep neural networks for such
tasks.

\subsection*{Document Revision History}
\label{sec:changelog}

This revision contains updated experimental results using a TensorFlow
implementation of SparseNet.  These results differ slightly from those reported
earlier and those in the version of this work published at ECCV 2018.  The
discrepancy stems from a minor bug, which acted as a source of regularization
when training SparseNet[$\oplus$], in our earlier PyTorch~\cite{paszkepytorch}
implementation.  The TensorFlow implementation corrects this bug.  The updated
results, which remove extra regularization from SparseNet[$\oplus$], are a
fairer comparison to DenseNet; the overall parameter efficiency advantages of
SparseNet[$\oplus$] remain significant.

\bibliographystyle{splncs04}
\bibliography{sparsenet_arxiv}

\begin{thebibliography}{10}
\providecommand{\url}[1]{\texttt{#1}}
\providecommand{\urlprefix}{URL }
\providecommand{\doi}[1]{https://doi.org/#1}

\bibitem{tensorflow}
Abadi, M., Agarwal, A., Barham, P., Brevdo, E., Chen, Z., Citro, C., Corrado,
  G.S., Davis, A., Dean, J., Devin, M., Ghemawat, S., Goodfellow, I., Harp, A.,
  Irving, G., Isard, M., Jia, Y., Jozefowicz, R., Kaiser, L., Kudlur, M.,
  Levenberg, J., Man\'{e}, D., Monga, R., Moore, S., Murray, D., Olah, C.,
  Schuster, M., Shlens, J., Steiner, B., Sutskever, I., Talwar, K., Tucker, P.,
  Vanhoucke, V., Vasudevan, V., Vi\'{e}gas, F., Vinyals, O., Warden, P.,
  Wattenberg, M., Wicke, M., Yu, Y., Zheng, X.: {TensorFlow}: Large-scale
  machine learning on heterogeneous systems (2015),
  \url{http://tensorflow.org/}

\bibitem{SegNet}
Badrinarayanan, V., Kendall, A., Cipolla, R.: {SegNet}: A deep convolutional
  encoder-decoder architecture for image segmentation. PAMI  (2017)

\bibitem{RevArbDeep}
Chang, B., Meng, L., Haber, E., Ruthotto, L., Begert, D., Holtham, E.:
  Reversible architectures for arbitrarily deep residual neural networks. AAAI
  (2018)

\bibitem{DeepLab}
Chen, L.C., Papandreou, G., Kokkinos, I., Murphy, K., Yuille, A.L.: {DeepLab}:
  Semantic image segmentation with deep convolutional nets, atrous convolution,
  and fully connected {CRFs}. arXiv:1606.00915  (2016)

\bibitem{ChenWTWC15}
Chen, W., Wilson, J.T., Tyree, S., Weinberger, K.Q., Chen, Y.: Compressing
  neural networks with the hashing trick. ICML  (2015)

\bibitem{DualPath}
Chen, Y., Li, J., Xiao, H., Jin, X., Yan, S., Feng, J.: Dual path networks.
  NIPS  (2017)

\bibitem{deng2009imagenet}
Deng, J., Dong, W., Socher, R., Li, L.J., Li, K., Fei-Fei, L.: {ImageNet}: A
  large-scale hierarchical image database. CVPR  (2009)

\bibitem{girshick2014rich}
Girshick, R., Donahue, J., Darrell, T., Malik, J.: Rich feature hierarchies for
  accurate object detection and semantic segmentation. CVPR  (2014)

\bibitem{RevNet}
Gomez, A.N., Ren, M., Urtasun, R., Grosse, R.B.: The reversible residual
  network: Backpropagation without storing activations. NIPS  (2017)

\bibitem{gray2017gpu}
Gray, S., Radford, A., Kingma, D.P.: {GPU} kernels for block-sparse weights.
  Tech. rep., OpenAI (2017)

\bibitem{unrolled}
Greff, K., Srivastava, R.K., Schmidhuber, J.: Highway and residual networks
  learn unrolled iterative estimation. ICLR  (2017)

\bibitem{fb_resnet_torch}
Gross, S., Wilber, M.: Training and investigating residual nets  (2016),
  \url{https://github.com/facebook/fb.resnet.torch}

\bibitem{han2015deep_compress}
Han, S., Mao, H., Dally, W.J.: Deep compression: Compressing deep neural
  networks with pruning, trained quantization and huffman coding. ICLR  (2016)

\bibitem{HAGM:CVPR:2015}
Hariharan, B., Arbelaez, P., Girshick, R., Malik, J.: Hypercolumns for object
  segmentation and fine-grained localization. CVPR  (2015)

\bibitem{maskrcnn}
He, K., Gkioxari, G., Dollar, P., Girshick, R.: Mask {R-CNN}. ICCV  (2017)

\bibitem{he2015delving}
He, K., Zhang, X., Ren, S., Sun, J.: Delving deep into rectifiers: Surpassing
  human-level performance on {ImageNet} classification. ICCV  (2015)

\bibitem{he2016deep}
He, K., Zhang, X., Ren, S., Sun, J.: Deep residual learning for image
  recognition. CVPR  (2016)

\bibitem{he2016preact}
He, K., Zhang, X., Ren, S., Sun, J.: Identity mappings in deep residual
  networks. ECCV  (2016)

\bibitem{logDenseNet}
Hu, H., Dey, D., Giorno, A.D., Hebert, M., Bagnell, J.A.: {Log-DenseNet}: How
  to sparsify a {DenseNet}. arXiv:1711.00002  (2017)

\bibitem{CondenseNet}
Huang, G., Liu, S., van~der Maaten, L., Weinberger, K.Q.: {CondenseNet}: An
  efficient {DenseNet} using learned group convolutions. CVPR  (2018)

\bibitem{huang2016densely}
Huang, G., Liu, Z., van~der Maaten, L., Weinberger, K.Q.: Densely connected
  convolutional networks. CVPR  (2017)

\bibitem{huang2016deep}
Huang, G., Sun, Y., Liu, Z., Sedra, D., Weinberger, K.Q.: Deep networks with
  stochastic depth. ECCV  (2016)

\bibitem{SqueezeNet}
Iandola, F.N., Moskewicz, M.W., Ashraf, K., Han, S., Dally, W.J., Keutzer, K.:
  {SqueezeNet}: {AlexNet}-level accuracy with 50x fewer parameters and $<${1MB}
  model size. arXiv:1602.07360  (2016)

\bibitem{ioffe2015batch}
Ioffe, S., Szegedy, C.: Batch normalization: Accelerating deep network training
  by reducing internal covariate shift. ICML  (2015)

\bibitem{krizhevsky2009learning}
Krizhevsky, A., Hinton, G.: Learning multiple layers of features from tiny
  images. Tech. rep., University of Toronto (2009)

\bibitem{krizhevsky2012imagenet}
Krizhevsky, A., Sutskever, I., Hinton, G.E.: {ImageNet} classification with
  deep convolutional neural networks. NIPS  (2012)

\bibitem{FractalNet}
Larsson, G., Maire, M., Shakhnarovich, G.: {FractalNet}: Ultra-deep neural
  networks without residuals. ICLR  (2017)

\bibitem{lee2014deeply}
Lee, C.Y., Xie, S., Gallagher, P., Zhang, Z., Tu, Z.: Deeply-supervised nets.
  AISTATS  (2015)

\bibitem{FCN}
Long, J., Shelhamer, E., Darrell, T.: Fully convolutional networks for semantic
  segmentation. CVPR  (2015)

\bibitem{paszkepytorch}
Paszke, A., Chintala, S., Collobert, R., Kavukcuoglu, K., Farabet, C., Bengio,
  S., Melvin, I., Weston, J., Mariethoz, J.: {PyTorch}: Tensors and dynamic
  neural networks in python with strong {GPU} acceleration, {May} 2017

\bibitem{DeepExpander}
Prabhu, A., Varma, G., Namboodiri, A.M.: Deep expander networks: Efficient deep
  networks from graph theory. arXiv:1711.08757  (2017)

\bibitem{UNet}
Ronneberger, O., Fischer, P., Brox, T.: {U-Net}: Convolutional networks for
  biomedical image segmentation. MICCAI  (2015)

\bibitem{simonyan2014very}
Simonyan, K., Zisserman, A.: Very deep convolutional networks for large-scale
  image recognition. ICLR  (2015)

\bibitem{srivastava2015highway}
Srivastava, R.K., Greff, K., Schmidhuber, J.: Highway networks.
  arXiv:1505.00387  (2015)

\bibitem{szegedy2017inception}
Szegedy, C., Ioffe, S., Vanhoucke, V., Alemi, A.A.: {Inception-v4},
  {Inception-ResNet} and the impact of residual connections on learning. AAAI
  (2017)

\bibitem{szegedy2015going}
Szegedy, C., Liu, W., Jia, Y., Sermanet, P., Reed, S., Anguelov, D., Erhan, D.,
  Vanhoucke, V., Rabinovich, A.: Going deeper with convolutions. CVPR  (2015)

\bibitem{MixedLink}
Wang, W., Li, X., Yang, J., Lu, T.: Mixed link networks. arXiv:1802.01808
  (2018)

\bibitem{LearnStructSparse}
Wen, W., Wu, C., Wang, Y., Chen, Y., Li, H.: Learning structured sparsity in
  deep neural networks. NIPS  (2016)

\bibitem{DynamicalIsometry}
Xiao, L., Bahri, Y., Sohl-Dickstein, J., Schoenholz, S.S., Pennington, J.:
  Dynamical isometry and a mean field theory of {CNNs}: How to train
  10,000-layer vanilla convolutional neural networks. ICML  (2018)

\bibitem{xie2016aggregated}
Xie, S., Girshick, R., Doll{\'a}r, P., Tu, Z., He, K.: Aggregated residual
  transformations for deep neural networks. CVPR  (2017)

\bibitem{sergey2016wideresnet}
Zagoruyko, S., Komodakis, N.: Wide residual networks. BMVC  (2016)

\bibitem{PSPNet}
Zhao, H., Shi, J., Qi, X., Wang, X., Jia, J.: Pyramid scene parsing network.
  CVPR  (2017)

\bibitem{NAS}
Zoph, B., Le, Q.V.: Neural architecture search with reinforcement learning.
  arXiv:1611.01578  (2016)

\end{thebibliography}
\end{document}